# A Cooperative Perception Environment for Traffic Operations and Control

**Hanlin Chen**
Lyles School of Civil Engineering, Purdue University
550 Stadium Mall Drive, West Lafayette, IN 47907
Email: chen1368@purdue.edu

**Brian Liu**
Saline High School
1300 Campus Pkwy, Saline, MI 48176
Email: liub@salinehornets.org

**Xumiao Zhang**
Department of Electrical Engineering and Computer Science, University of Michigan
2260 Hayward Street, Ann Arbor, MI 48109
Email: xumiao@umich.edu

**Feng Qian**
Department of Computer Science and Engineering, University Minnesota, Twin Cities
Kenneth H. Keller Hall, Minneapolis, MN 55455
Email:  fengqian@umn.edu

**Z. Morley Mao**
Department of Electrical Engineering and Computer Science, University of Michigan
2260 Hayward Street, Ann Arbor, MI 48109
Email: zmao@umich.edu

**Yiheng Feng, Corresponding Author**
Lyles School of Civil Engineering, Purdue University
550 Stadium Mall Drive, West Lafayette, IN 47907
Email: feng333@purdue.edu

**ABSTRACT**
Existing data collection methods for traffic operations and control usually rely on infrastructure-based loop detectors or probe vehicle trajectories. Connected and automated vehicles (CAVs) not only can report data about themselves but also can provide the status of all detected surrounding vehicles. Integration of perception data from multiple CAVs as well as infrastructure sensors (e.g., LiDAR) can provide richer information even under a very low penetration rate. This paper aims to develop a cooperative data collection system, which integrates Lidar point cloud data from both infrastructure and CAVs to create a cooperative perception environment for various transportation applications. The state-of-the-art 3D detection models are applied to detect vehicles in the merged point cloud. We test the proposed cooperative perception environment with the max pressure adaptive signal control model in a co-simulation platform with CARLA and SUMO. Results show that very low penetration rates of CAV plus an infrastructure sensor are sufficient to achieve comparable performance with 30% or higher penetration rates of connected vehicles (CV). We also show the equivalent CV penetration rate (E-CVPR) under different CAV penetration rates to demonstrate the data collection efficiency of the cooperative perception environment.

## INTRODUCTION

Traffic operations and control applications (e.g., actuated/adaptive traffic signal control) require real-time traffic information. Traditional infrastructure-based sensor systems such as loop-detectors and traffic cameras have been widely implemented in the field for decades. Infrastructure-based sense systems usually have relatively high installation and maintenance costs. More importantly, data collected from traditional infrastructure-based sensors is location-specific, which does not reflect the whole spatial distribution of vehicles. In the past decade, with the emergence of connected vehicle (CV) technology, CV data shows significant advantages over infrastructure-based sensor data because it contains continuous spatiotemporal trajectories, which provide much more traffic information. Trajectory-based traffic estimation and control methods have been developed and utilized in various applications (*1*). However, a certain penetration rate is required to receive benefits from CV data. For real-time applications, the critical penetration rate is usually around 20%-30% (*2*) (*3*). It may take a long time to reach such a penetration rate, especially given that the vehicle-to-vehicle (V2V) communication mandate was dropped in the U.S.

Connected and Autonomous vehicles (CAVs) perceive the driving environment with onboard perception sensors such as LiDAR and camera. The perception sensors not only can be used to provide data for CAV navigation but also a potential new data source for traffic operations and control. Compared with CV trajectories, perception data from CAV has the following two major advantages. 1) CV trajectories only report the status of the ego vehicles, while the perception data from CAVs also include the status of all detected surrounding vehicles within the sensor range. This means that the required penetration rate for CAV to provide a similar level of data would be much lower than CV. For example, a recent study shows that less than 10% of CAV could collect more than 50% of the total trajectory points in the traffic (*4*). 2) CV trajectories usually have relatively large position errors due to commercial grade GPS compared to CAVs, which may significantly impact applications requiring a higher level of accuracy (e.g., lane level mapping). On the other hand, the GPS devices installed on autonomous vehicles usually have much higher resolutions (e.g., with RTK corrections). With the help of other sensors through multi-sensor fusion (MSF), the localization accuracy of a CAV can reach centimeter level *(5)*. The accuracy of localization results can improve with MSF even in driving environments where GPS signals are not reliable *(6)*. Meanwhile, onboard perception sensors such as LiDAR also have centimeter-level accuracy. As a result, the generated surrounding vehicle trajectories from CAVs would be much more precise than those from CVs.

Besides CAVs, future digital infrastructure is also expected to be installed with perception sensors and communication systems similar to CAVs to further enhance the traffic monitoring capabilities and enable cooperative driving automation (CDA). The Society of Automotive Engineers (SAE) defines four classes of CDA in the J3216 standard, including status-sharing, intent-sharing, agreement-seeking, and prescriptive (*7*). Under different cooperation classes, different CDA applications can be developed. To implement CDA applications or to maximize the data utilization from multiple CAVs for traffic operations and control, a cooperative perception environment is indispensable, which has become a trending research topic in the past few years. However, existing studies usually simplify this problem by assuming all vehicles in the detection range (e.g., 100m) can be detected by the CAV without errors (*8*) or applying simple geometric models to exclude occluded vehicles (*4*). In reality, 2D/3D detection models do not have 100% accuracy. The detection precision is associated with many factors, including but not limited to road geometry (road grade and curvature), sensor configurations, operating conditions (weather and ambient light), and types of detection model (deep-learning based or none deep-learning based). It is necessary to investigate the performance of the detection models under different driving environments and analyze their impact on downstream CDA and traffic operation applications.

This paper aims to develop such a framework, which integrates Lidar sensor data from both infrastructure and CAVs to create a cooperative perception environment for various transportation applications. The framework is developed based on a multi-resolution simulation platform consisting of CARLA and SUMO. CARLA is the most widely adopted open-source simulator for autonomous driving systems, and SUMO is a microscopic simulator for traffic simulation. Lidars are mounted on the CAVs and at the intersections in the CARLA environment to generate simulated point cloud data. The point cloud data from different entities are first merged to enhance the perception quality, and then state-of-the-art 3D Lidar detection models are applied to detect vehicles. After geofencing and redundancy reduction of the detection results, vehicle trajectories can be extracted, which serve as inputs to the downstream applications. We evaluate the proposed cooperative perception framework with the adaptive traffic signal control application, where the max-pressure (*9*) algorithm is applied. In addition, we compare the performance of the cooperative perception-based traffic control with the trajectory-based control where trajectories are reported by CVs. To the best of our knowledge, this is the first study that integrates a full Lidar-based cooperative sensing and perception pipeline with traffic operation and control applications.

The rest of the paper is organized as follows. Section 2 will introduce the cooperative perception environment in detail. Section 3 focuses on how this environment can be integrated with adaptive traffic signal control and presents the results and findings. Section 4 provides discussions on the advantages and disadvantages of the proposed framework and lays out some future research directions. Finally, Section 5 concludes the paper.

## COOPERATIVE PERCEPTION ENVIRONMENT
### System Overview
Figure 1 shows the overall structure of the proposed cooperative perception environment. We develop the system based on a multi-resolution co-simulation environment consisting of CARLA and SUMO. Lidar sensors are deployed on CAVs and infrastructure in CARLA and generate point cloud data. Then raw data level merging is performed to merge point clouds from multiple Lidars. The merged point clouds are fed into a 3D object detection model to obtain detected vehicle locations (i.e., bounding boxes). Then a decision-level data fusion is performed with road geometry information to remove noises and redundant detections. After data filtering, the generated vehicle information (e.g., location) is sent to downstream applications. The output of the application (e.g., traffic signal plan, optimized vehicle trajectories) will be sent back to the co-simulation environment for execution. In the next few subsections, we will elaborate on the co-simulation platform, raw data level point cloud merging, 3D object detection model, and the decision level data merging.

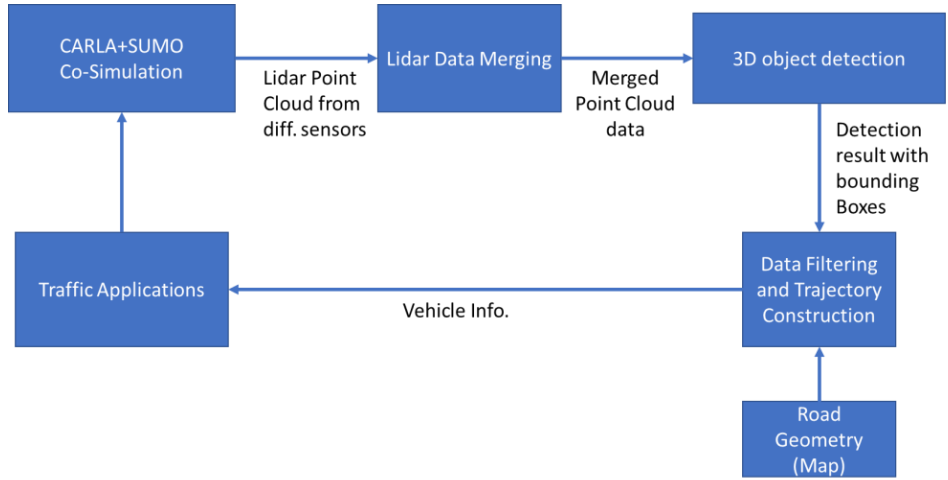

**Figure 1:** Overview of the Cooperative Perception Environment

**Co-Simulation Environment**

We implement the proposed cooperative perception system in an open-source co-simulation platform developed by the CARMA program at Federal Highway Administration (*10*). The co-simulation platform integrates simulation tools, namely SUMO and CARLA. SUMO is a microscopic traffic simulator that is used to simulate realistic traffic flow and traffic control logic. CARLA provides vehicle-level simulation, including but not limited to vehicle dynamics and sensor simulation. LiDAR data provided by CARLA includes both location information and reflection rates, which have been utilized extensively in deep-learning-based 3D object detection algorithms (*11*, *12*).

The co-simulation platform utilizes Mosaic as the central controller to synchronize data between different simulation platforms. Vehicles can be generated in either SUMO or CARLA, and the corresponding vehicle status such position will be synchronized to the other simulator through Mosaic. The same mechanism is applied to traffic signal status. Several traffic networks and traffic scenarios have been built and tested in the co-simulation tool and are available to the users. The co-simulation platform is illustrated in Figure 2, where the left part is the CARLA view, and the right part is the SUMO view. Based on the co-simulation platform, we add several new functions to implement the cooperative perception environment, as detailed in the following sections.

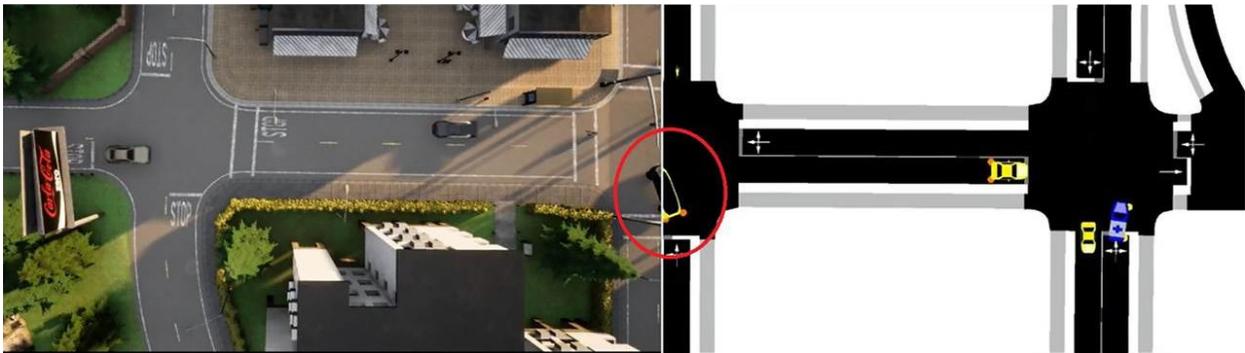

**Figure 2:** Co-Simulation Platform from the CARMA Program
(Source: https://usdot-carma.atlassian.net/wiki/spaces/CRMSIM/overview)

**Lidar Data Processing**

*Raw Lidar Data Merging*

General cooperative perception has three layers of data sharing and merging: raw data level, feature level, and decision level (*13*). Raw data level merging ensures minimum information loss but can only be applied when two data sources share the same data structures. Meanwhile, raw data sharing usually requires the highest communication bandwidth, which may cause high latency in the transmission. Decision-level data merging is comparatively the easiest to implement but may suffer from information loss in previous data processing stages. Feature level fusion usually appears in multi-sensor fusion from different sensors. In this work, we integrate both raw data level merging and decision level data merging. The benefits of the cooperative perception scheme are significant. When considering a point cloud obtained from a single lidar, miss detection can occur during the process of 3D object detection. Apart from the detection model itself, missing (e.g., due to occlusion or limited range of field of view), sparse (e.g., due to long distance), or different patterns of point clouds can all contribute to the miss detection. Given the limitations in single Lidar detection, raw-data level merging has been proposed to reduce data missing in the single Lidar point cloud. In this paper, we apply an algorithm called EMP (Edge-Assisted Multi-vehicle Perception) (*14*) to merge point cloud data from multiple Lidar sensors. The merging algorithm is briefly introduced below.

For all Lidar devices that appeared in the system, we first obtain their location and rotation information in terms of x, y, z, yaw, pitch, and roll. Second, we obtain the height of LiDAR with respect to the ground. We select one Lidar as the ego device (could be either from a CAV or infrastructure) and mark the corresponding point cloud data and location data as primary information. The location and point cloud data from other Lidars are grouped as secondary information. Then we perform a coordinate transformation with regard to the ego device's location. Rotation matrix R is obtained using differences in yaw, pitch, and roll of the two devices, where $d_{yaw}, d_{pitch}, and\ d_{roll}$ refer to the angle difference in yaw, pitch, and roll. The rotation matrix is expressed as:

$$R_z = \begin{bmatrix} cos(d_{yaw}) & -sin(d_{yaw}) & 0 & 0 \\ sin(d_{yaw}) & cos(d_{yaw}) & 0 & 0 \\ 0 & 0 & 1 & 0 \\ 0 & 0 & 0 & 1 \end{bmatrix} \qquad \text{Equation 1}$$

$$R_y = \begin{bmatrix} cos(d_{yaw}) & 0 & sin(d_{yaw}) & 0 \\ 0 & 1 & 0 & 0 \\ -sin(d_{yaw}) & 0 & cos(d_{yaw}) & 0 \\ 0 & 0 & 0 & 1 \end{bmatrix} \qquad \text{Equation 2}$$

$$R_x = \begin{bmatrix} 1 & 0 & 0 & 0 \\ 0 & cos(d_{roll}) & -sin(d_{roll}) & 0 \\ 0 & sind_{roll} & cos(d_{roll}) & 0 \\ 0 & 0 & 0 & 1 \end{bmatrix} \qquad \text{Equation 3}$$

$$R = R_z R_y R_x \qquad \text{Equation 4}$$

For each pair of point cloud conversion, a translation matrix $T$ is defined by the following equations, where $\Delta x$, $\Delta y$, $and\ \Delta z$ are the differences between the ego location and secondary location in *x, y,* and *z* coordinates, respectively. Note that the calculation of z coordinates also includes the mounting height of LiDAR sensors, for which we use $height_{diff}$ as the height difference between sensor mounting points:

$$dx = \Delta x * cos(-Yaw_{ego}) + \Delta y * sin(-Yaw_{ego})$$

Equation 5

$$dy = -\left(\Delta x * -sin(-Yaw_{ego}) + \Delta y * cos(-Yaw_{ego})\right)$$

Equation 6

$$dz = \Delta z + height_{diff}$$

Equation 7

$$T = [dx \quad dy \quad dz \quad 0]$$

Equation 8

Algorithm 1 describes the raw Lidar data merging process.

**Algorithm: Raw Lidar Data Merging**

**Input:** Set of point cloud from all Lidar sensors $\boldsymbol{P}_i$, with sensor locations $L_i = \{x_i, y_i, z_i\}$, rotations $RO_i = \{yaw_i, pitch_i, roll_i\}$, sensor heights $H_i$, and $i \in \boldsymbol{I}$;

**Begin**

    Select an ego Lidar sensor in $\boldsymbol{I}$ with location $L_0$, rotation $RO_0$ sensor height $H_0$, and point cloud data $\boldsymbol{P}_0$

    **For** $i$ in $\boldsymbol{I} \setminus \{i_0\}$ **do**

        Obtain $P_i, L_i, RO_i$, and $H_i$

        Calculate rotation matrix $R_{0-i}$ via Equations 1-4

        Calculate translation matrix $T_{0-i}$ via Equations 5-8

        **For** $j$ in $\boldsymbol{P}_i$ **do**

            $P_i^j = R_{0-i}P_i + T_{0-i}$

            Append $P_i^j$ to $P_0$

        **End For**

    **End For**

**End**

**Output:** Merged point cloud from all Lidar sensors ($P_0$)

Where $\boldsymbol{I}$ is the set of Lidar sensors in the network, and $P_i^j$ is the $j^{\text{th}}$ point in $i^{\text{th}}$ Lidar point cloud data.

Figure 3 shows an example of the merged point cloud. Figure 3a, 3b, and 3c represent Lidar point clouds collected from an infrastructure Lidar, a Lidar mounted on a CAV going westbound, and a Lidar mounted on a CAV going eastbound, respectively. Figure 3d shows the merged Lidar point clouds. The merged point cloud data is then fed into 3D Lidar detection models to obtain information of surrounding vehicles, as introduced in the next section.

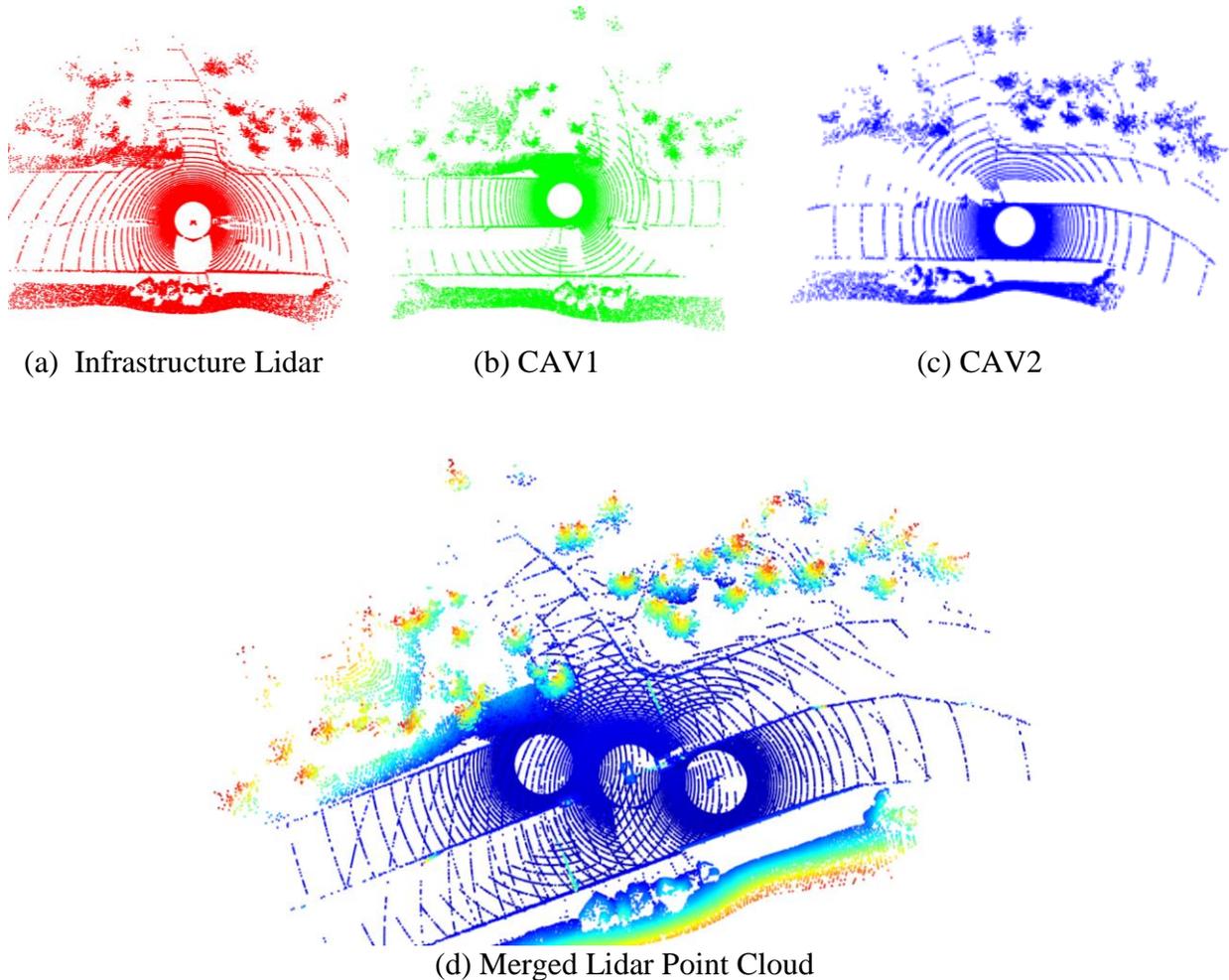

(a) Infrastructure Lidar                (b) CAV1                      (c) CAV2

(d) Merged Lidar Point Cloud
**Figure 3:** Raw Lidar Data Merging

### 3D Lidar detection model

Two 3D Lidar detection models are introduced in this section. The first one is a none deep learning based model, DBSCAN. The second one is a deep learning based model, PointPillar. Their performances at the detection level and application level will be compared in the experiment section.

### Clustering-Based 3D Detection Model

We implement the Density-Based Algorithm for Discovering Clusters in Large Spatial Databases with Noise (DBSCAN)(*15*) as the clustering-based detector for detecting the surrounding vehicle's location. Similar to the pure Euclidean-distance-based clustering method, DBSCAN is also a distance-related clustering algorithm. Instead of considering the distance between points only, it considers the density of points distribution and includes points that have a higher density than others. As the performance of the DBSCAN clustering algorithm highly depends on the result of ground removal, we first implement the Random sample consensus (RANSAC) (*16*) algorithm to remove the point cloud data on the ground plane, which introduces noises in the detection. RANSAC is a method to estimate a plane that can include the maximum possible number of data points given a dataset. The output of RANSAC is the mathematical representation of the majority of points within the dataset, and the rest of the points are regarded as outliers. In the case of 3D point cloud data, we can assume that most of the points are from the ground plane. Therefore, we can use RANSAC to implement ground removal and keep the detected outliers as non-ground points. To ensure the maximum ground removal effect while keeping the other point data relevant to vehicles as much as possible, we perform RANSAC twice with a low distance threshold. This adjustment also ensures performance when the ground area is uneven within the detection range. For each round of RANSAC, we choose the distance threshold as 0.2 meters, the number of points taken into consideration within the iteration is 3, and the maximum num of iterations is 3000.

After ground removal, we perform the DBSCAN algorithm with the eps-neighborhood distance as 1.25 meters and the minimum number of points considered within a cluster as 3. To ensure the performance of the downstream applications and decrease the impact of detection results that are not vehicles, we perform size-based filtering based on the results from DBSCAN. We only consider clusters as vehicles if they have a length between 0.5 m and 6 m, widths between 0.5m and 3m, and heights between 0.1m and 2m. Points that belong to the ground plane can have similar densities and, therefore, will be regarded as an accepted clustering result by DBSCAN since the size of the cluster can be smaller than the upper bound. Thus, we add a lower bound to decrease the impact of such noisy predictions.

*Deep Learning Based 3D Detection Model.*
PointPillar (*11*) is chosen as the deep learning-based detection model because of its good detection accuracy relative to other popular detection models in combination with a fast inference speed, reported at 62 Hz compared to other models such as VoxelNet (*17*), which runs at only 4.4 Hz. Derived from the SECOND detection model (*12*), PointPillar employs an innovative point cloud encoder that uses vertical pillars instead of 3D cubes to represent voxels. The use of pillars allows PointPillar to only use 2D convolutions instead of having to use both 2D and 3D convolutions. 3D convolutions had been a major bottleneck for the previous models that PointPillar is based on, such as SECOND and VoxelNet. Eliminating them allows PointPillar to achieve high inference speed while being deployed in real-time. Figure 4 provides an overview of the PointPillar network structure. The Pillar Feature Net transforms the raw point cloud data into stacked pillars that are then scattered back to a pseudo image. The backbone then does feature extraction on the pseudo image, and the detection head uses those features to come up with its predictions.

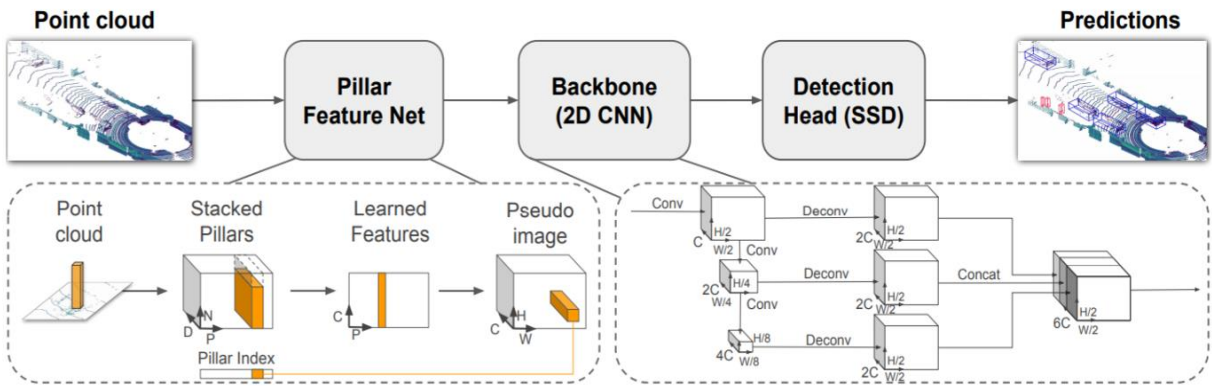

**Figure 4:** PointPillar Network Structure (*11*)

In our implementation, the Point Pillar's parameters differ from the default values due to differences between the CARLA simulation and the KITTI dataset that the PointPillar model was originally configured for. The model needs to have a 360° inference to accommodate the needs of downstream applications instead of 180° in the originally KITTI-trained model. As a result, the range of the model is extended to (-80, 80) on both the *x* and *y* axis. The dimensions of the pseudo-image created from the PointPillar scatter are extended to (640, 640) and the voxel size is extended to (0.25, 0.25, 4). This is to compensate for the doubled detection range. Testing multiple PointPillar model configurations found that this voxel size is optimal for this study. Finally, the car is the only object class for this model, as the simulation contains no pedestrians and cyclists, and the car anchor is changed to (1.7, 4.4, 1.7) to match the vehicle size in the CARLA simulation environment.

*Decision-level data merging*
With the 3D lidar detection model described in the previous section, we obtain the locations of surrounding vehicles in the local coordinates of each Lidar. Note that the 3D lidar detection model is trained with a single-point cloud only. Multiple inferences have to be made for each CAV and infrastructure Lidar. Then we obtain detection results from each Lidar and convert them to a global coordinate. A decision-level data merging is performed to combine all the detected surrounding vehicles' information. Given the same object can be detected multiple times by different Lidars when they have overlapping point clouds, which may have an impact on the performance of downstream applications, a redundancy check, and removal mechanism are needed. In this study, we compare the locations of detected vehicles. If the distance between two detections is smaller than a threshold (e.g., one vehicle length), we consider they belong to the same vehicle and delete the one with a longer distance from the intersection. In addition to redundancy removal, geofencing is performed to remove all detected objects that are outside of the road geometry. Finally, a lane mapping algorithm is performed to match the detected vehicles to the lane level and calculate their distances to the stop bar. After removing the redundant detections and geofencing, the final results are sent to the downstream traffic applications. Figure 5 shows a comparison between with and without raw and decision level Lidar data merging. Figure 5(a) shows the detection results based on a single CAV Lidar and Figure 5(b) shows the results based on a merged Lidar point cloud from the CAV and the infrastructure Lidar. It can be seen that a few more vehicles that are not detected before can be detected after merging.

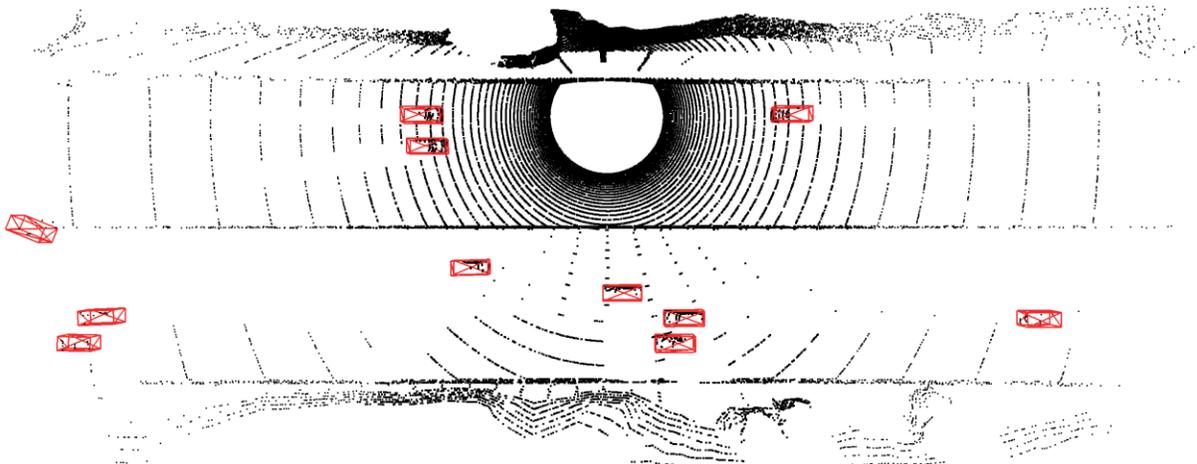

(a): detection results based on single Lidar point cloud

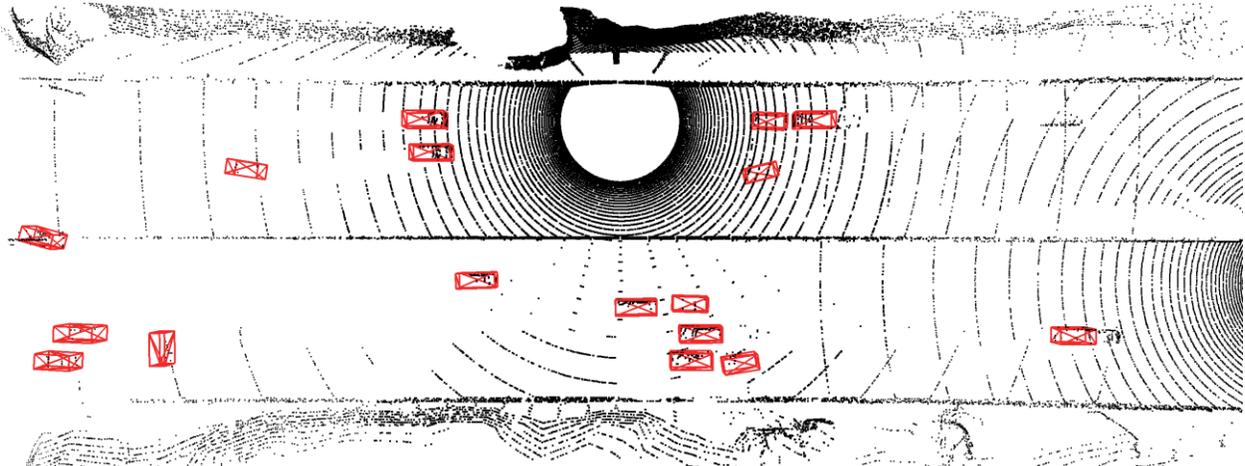

(b): detection result based on merged Lidar point cloud

**Figure 5:** Detection Results Comparison Before and After Raw and Decision Level Merging

## A CASE STUDY – ADAPTIVE TRAFFIC SIGNAL CONTROL

### Experiment Setting

To demonstrate the proposed cooperative perception framework for traffic operations and control applications, we conduct a case study on adaptive traffic signal control. The selected intersection is located in Town04, a virtual map provided by CARLA, as shown in Figure 6. It is a T-intersection with only two signal phases. Traffic is generated from SUMO, with traffic volumes on the main street to be 500 veh/hr/ln and volume on the side street to be 360 veh/h/lane. Two types of vehicles are assigned, non-CAV (which we refer to as HV) and CAV. The CAV penetration rate varies from 0% to 5%. For each generated CAV, a 64-layer LiDAR is attached with a height of 2.4m in relation to its center point. An infrastructure Lidar is installed at the center of the intersection. The Lidar installed at the intersection has a height of 3 meters. We run the co-simulation on Ubuntu 20.04 with an NVIDIA RTX 3090 GPU, 64GB RAM, and 66GB for swap files (a total of 120GB for temporary memory).

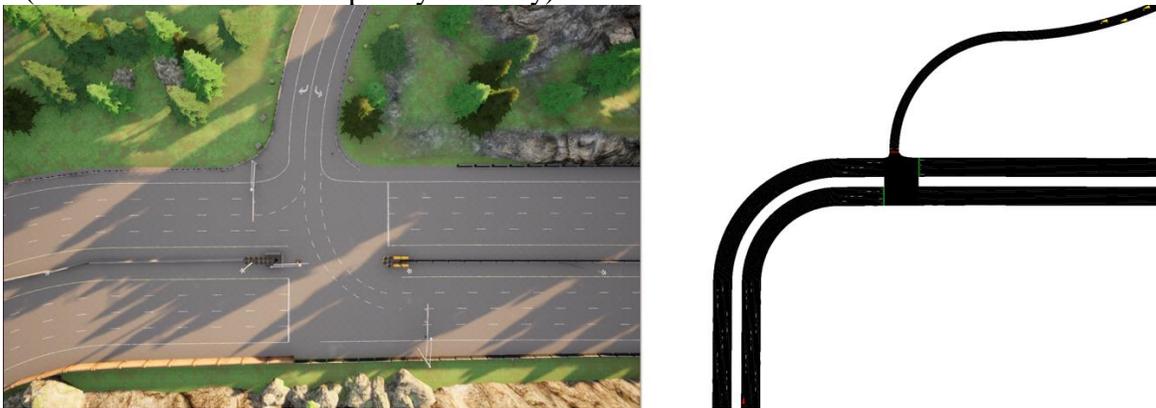

**Figure 6:** A Signalized Intersection in CARLA Town04 (Study Area)

(The left-hand side is the BEV of the intersection in CARLA, and the right-hand side is the same BEV from SUMO)

Max pressure (*9*) is applied as the adaptive signal control model. Max pressure is a well-known signal control strategy that calculates the pressure to determine the signal phasing. The pressure is defined as the difference between the number of incoming vehicles and departing vehicles upstream and downstream of the intersection, respectively. The signal phase that has the highest pressure will receive green time. The original max pressure model switches phases instantly without considering transition and minimum green time. Four seconds of yellow internal and 1 second of an all-red interval are added to make the model more practical. In addition, the minimum green time of the main street and the side street is set to 5 seconds. We choose max pressure as a proof of concept. Other signal control models can also be integrated with the system and will be left for future study. The input of the max pressure control (i.e., the number of vehicles on each signal phase) can be directly obtained from the processed detection results.

    We aim to evaluate the performance of the cooperative perception framework from three different perspectives. First, it is important to evaluate the performance of the 3D detection models, in which the precision metrics are presented and compared between two different types of detection models. Second, the detection results from Lidars under different CAV penetration rates are converted to equivalent connected vehicle penetration rate (E-CVPR), from which we can compare the data collection efficiency. Finally, the performance of the max pressure control in terms of average vehicle delay is compared between CV-based data as the input and cooperative perception-based data as the input. The following sections elaborate on the experiment results and analysis.

## Results and Analysis
### 3D Lidar Detection Model
We compare the performance of clustering-based and deep-learning-based 3D object detection models at the detection level. The evaluation metrics we use are defined below, following the evaluation matrix defined in the KITTI dataset (*18*). We first define the threshold for determining the true positive prediction. To evaluate the prediction result compared to ground truth, we calculate the intersection over union (IoU) score for each detection result, as defined in Equation 9. Then based on the IoU score, we classify prediction results as true positive, false positive, and false negative. Based on the classified prediction results, we obtain a precision-recall curve. Then we calculate the interpolated average precision score based on the curve obtained. In our work, we calculate a 40-point interpolated average precision score (*19*), denoted as $AP_{40}$. For calculating the 40-point interpolated average precision score, we first obtain evenly-spaced recall levels $R_{40} = \{1/40, 2/40, \ldots, 39/40, 1\}$, then for each recall level, we calculate the interpolated precision, defined in Equation 10. The $AP_{40}$ score is then obtained as the average of all of the interpolated precision scores, as defined in Equation 11.

$$IoU_{(A,B)} = \frac{A \cap B}{A \cup B}$$

Equation 9

$$\rho_{intercept}(r) = \max_{r':r' \geq r} \rho(r')$$

Equation 10

$$AP_{40} = \frac{1}{40} \sum_{r \in R_{40}} \rho_{intercept}(r)$$

Equation 11

    We evaluate the detection accuracy from both bird-eye-view (BEV) and overlapping 3D bounding boxes (3D) perspectives. We use 0.1 and 0.01 as the IOU threshold for true positive prediction. The results are shown in Table 1. An example is provided in Figure 7, which shows the detection results from two models in the same frame. The green bounding boxes are the ground

truth while the red bounding boxes are the detection results. It can be seen that the precision of the deep learning-based model (i.e., PointPillar) outperforms the clustering-based model by a great margin in all cases. There are not many discrepancies in the PointPillar with different views and IOUs, but these parameters have a significant impact on the DBSCAN model. The low precision in DBSCAN results from the high false positive predictions outside the road boundary, as shown in Figure 7(a). After geofencing, the results are very much improved. Also, redundancy reduction is implemented for downstream applications on the decision level merging. The redundancy mechanism merged multiple clustering results, giving the same vehicle as one on the application side. This also reduced the impact of redundant detection on the downstream application side.

Note that the IOU value thresholds we use in this paper are significantly different from the common value (e.g., 0.5 or 0.7) used in the computer vision community. The reason is that the target applications of the proposed platform are mainly for traffic operations and control, which require lower precision on vehicle locations (i.e., a high IOU) as long as the object (the vehicle) can be detected and tracked. For example, in the max pressure control application implemented in this paper, only the number of vehicles and their approach (lane) are needed. In addition, the precision values from the BEV are more critical since most traffic applications only consider traffic flow in a two-dimensional space.

**Table 1:** Precision Evaluation for Different Detection Models

|            | $AP40_{0.01}$ BEV | $AP40_{0.01}$ 3D | $AP40_{0.1}$ BEV | $AP40_{0.1}$ 3D |
|------------|-------------------|------------------|------------------|-----------------|
| DBSCAN     | 23.3318           | 16.7142          | 15.1852          | 7.8422          |
| PointPillar| 84.9086           | 84.9144          | 84.8281          | 84.8151         |

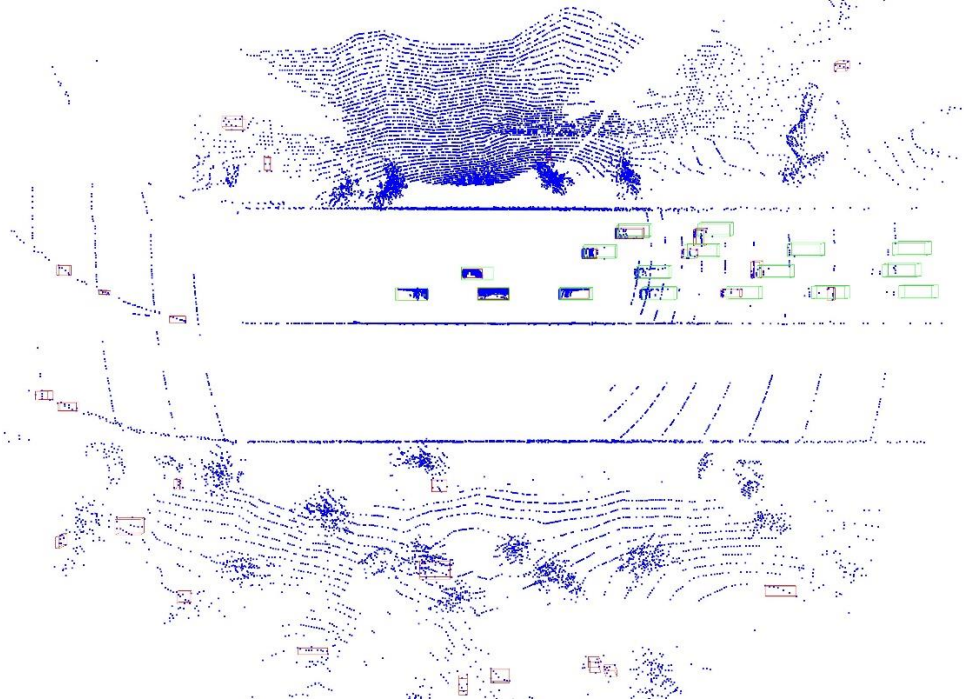

(a) detection result from none deep learning based detection model

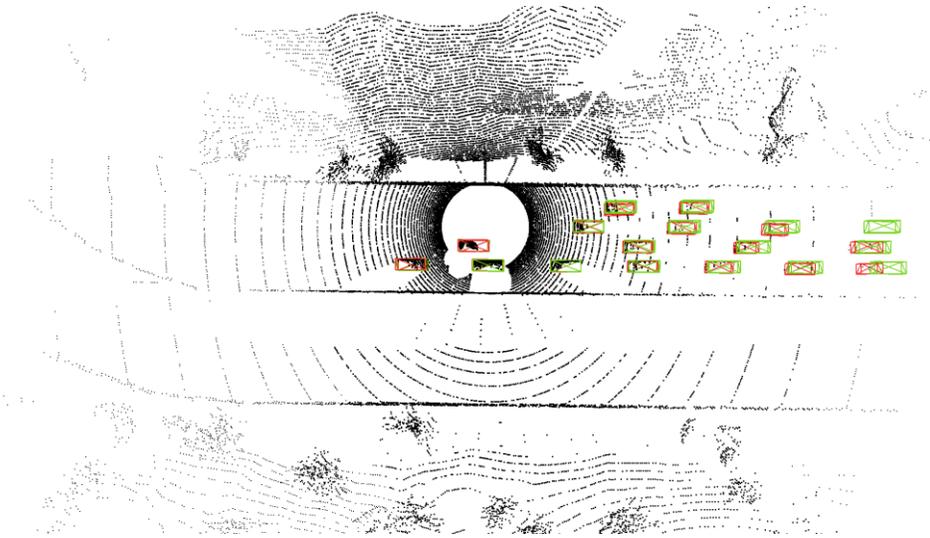

(b) detection result from deep learning based detection model

**Figure 7:** Detection Results Comparison

*Equivalent CV Penetration Rate*

One major benefit of the cooperative perception framework is the data collection efficiency. With a much lower CAV penetration rate, the observed vehicles would be equivalent to a much higher CV penetration rate. We define the equivalent CV penetration rate (E-CVPR) as the number of detected vehicles in the cooperative perception framework divided by the total number of vehicles on the roadway for a particular application. However, this ratio may vary under different traffic conditions, road geometries, vehicle demands, and infrastructure sensor locations. For example, road curvature and grade may not have a significant influence on the CV data, but it may impact the Lidar detection results adversely. As a result, measuring E-CVPR at different road segments may generate different results. In this study, we measure the E-CVPR within 200 meters upstream of the intersection and 100 meters downstream of the intersection. The reason we choose this area is that 1) it contains the infrastructure sensor so that even without any CAVs, a part of the vehicles can be detected by the infrastructure; and 2) the same area is used in the pressure calculation in max pressure control. Therefore, we emphasize that the E-CVPR value is application specific. Note that when calculating E-CVPR, only true detections are included, meaning that false positives are removed from the detection results. Table 2 shows the mean and standard deviation of E-CVPRs under different CAV penetration rates and different 3D-detection models. For each CAV penetration rate, there is always an infrastructure Lidar located in the middle of the intersection. That's why under 0% CAV penetration rate, the E-CVPR is still around 30%. Under a 5% CAV penetration rate, the E-CVPR can reach near 50%. Another finding is that although the precision value of the DBSCAN detector is much lower than the PointPillar detector, the E-CVPR is similar. This indicates that most of the misdetections from DBSCAN are located outside the roadway. After the decision level data merging, most of the noises are removed.

**Table 2:** Average E-CVPR under Different CAV Penetration Rates and Different 3D-Detection Models

| 3D Detection Model | 0% CAV PR | 1% CAV PR | 2% CAV PR | 5% CAV PR |
|---|---|---|---|---|
| DBSCAN | 29.88% | 35.43% | 38.26% | -- |

| | (10.74%) | (14.44%) | (16.69%) | |
|---|---|---|---|---|
| PointPillar | 31.4155% (10.12%) | 35.6826% (10.94%) | 37.4261% (11.89%) | 46.5995% (14.64%) |

Figure 8 shows a histogram of the E-CVPR under 5% of CAVs with PointPillar as the detection model. The variation of the E-CVPR is mainly caused by the uncertainties of the CAV arrivals at the intersection and the coverage areas by multiple Lidars.

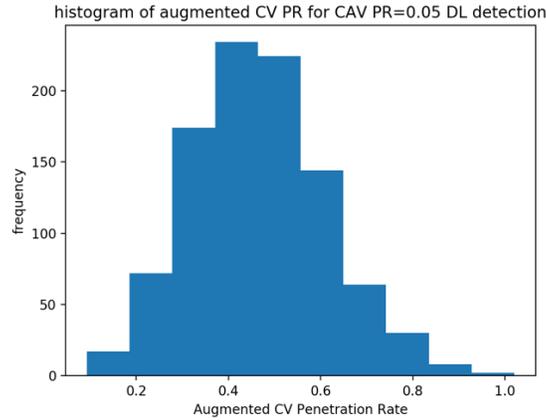

**Figure 8:** Histogram of E-CVPR for 5% CAV Penetration Rate with PointPillar Model

*Max Pressure Control*
Finally, we show the performance of the max pressure control under the cooperative perception framework. The results are also compared with the same control model in the CV environment. Table 3 shows the results from the CAV environment. At the application level, all penetration rates and detection models perform similarly, with an average delay between 10-11 seconds.

**Table 3 Average Vehicle Delay under different CAV penetration rates and different detection models**

| 3D Detection Model | 0% CAV PR | 1% CAV PR | 2% CAV PR | 5% CAV PR |
|---|---|---|---|---|
| DBSCAN | 10.98 | 10.99 | 10.86 | --- |
| PointPillar | 10.57 | 10.86 | 10.63 | 11.01 |

Figure 9 shows the average vehicle delay of the max pressure control under different CV penetration rates. It can be seen that the delay decreases with the increase in penetration rate, which meets our expectations. The performance of maximum pressure in the CAV environment fluctuates around 10 to 11 seconds when the penetration rate is 30% or higher in the CV environment, which is consistent with the results obtained in the CAV environment. In all cases of the CAV environment, the E-CVPR is around or higher than 30%.

Note that in Figure 9, the delay becomes stable after the penetration rate reaching to 30%. This is caused by the simplified settings of the intersection. The vehicle volume is quite low, and the traffic signal only contains two phases. For many cycles, the phase switches after the minimum green time due to changed pressure, when the number of departing vehicles is greater than the number of arriving vehicles. When the CV penetration rate is low, it is possible that the conflicting phase doesn't have any CVs, so the phase is not actuated. Unfortunately, we are not able to increase the traffic volume and/or implement the framework at a more complicated intersection due to limitations in computing resources, although a powerful desktop server is used. In the future, we will redesign the architecture of the co-simulation platform to enable parallel and cloud computing, which can support a larger network, more CAVs, and higher traffic volumes.

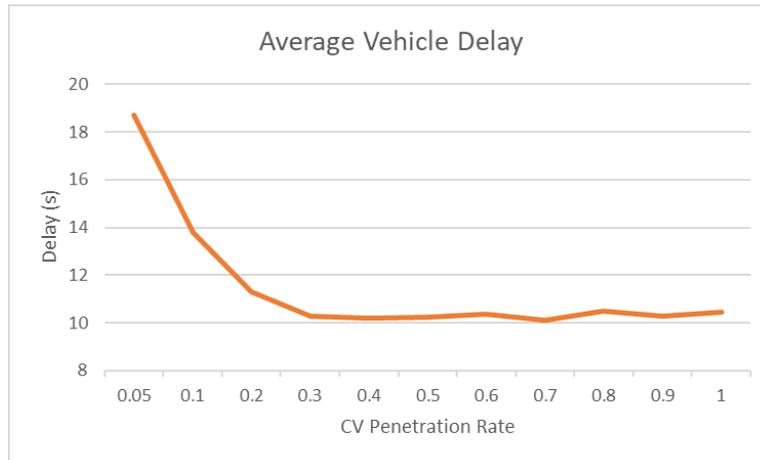

**Figure 9:** Average Vehicle Delay for Max Pressure Control in the CV Environment

## DISCUSSION

The proposed cooperative perception framework shows promising potential in data collection and implementation in traffic operations and control applications. There are still several key issues that need to be addressed in the future.

Currently, only vehicle location information is identified by the 3D detection model. This is sufficient for location-based applications such as max pressure control. For other applications where additional vehicle dynamic information is needed (e.g., speed, acceleration), object tracking needs to be incorporated. Moreover, when vehicles enter and exit the detection range of the CAVs, the ID of the vehicle will be changed. An association model needs to be developed to connect the same vehicle with different IDs. Although CVs are also supposed to change their IDs every few minutes, existing studies usually ignore this feature and assume each CV has a complete trajectory. The same assumption cannot be applied in the cooperative perception environment anymore.

Although the cooperative perception environment can greatly augment the penetration rate, still only partial traffic data are collected. The data pattern (e.g., trajectory spatial distribution) is significantly different from those collected in the CV environment. CVs are assumed to be distributed on the roadway randomly so that observed trajectories spread out the entire time and space domain. However, in the CAV environment, the distribution is highly skewed. The penetration rate within the detection ranges of CAVs or infrastructure sensors can be very high, while the penetration rate outside the detection ranges is essentially zero. Existing traffic state estimation models usually assume a CV environment. Therefore, new estimation models for the CAV environment are needed.

We advocate raw Lidar data fusion instead of later stage fusion because raw data level fusion can retain original information to the most extent. Examples in Figure 5 also show the benefits. However, merging raw Lidar data brings a new problem for the deep learning-based detection models. Existing models are trained with a single Lidar source by extracting features from the point cloud to match the label (e.g., a vehicle). After merging, the features that represent a vehicle may be different due to more Lidar points from different angles with different reflection values. This may bring challenges to the deep learning-based detection models. In our experiments, we do observe some cases where the vehicle can be detected before Lidar point cloud merging but gets lost after the merging. New detection models need to be trained based on the merged point cloud data.

## CONCLUSIONS

In this paper, we developed a Lidar-based cooperative perception environment for traffic operations and control applications. The environment was built based on a co-simulation platform consisting of CARLA and SUMO. Raw Lidar point cloud data from multiple sources (CAVs and infrastructure) were merged, and the state-of-the-art 3D Lidar detection models were applied to detect vehicles from the point cloud. The detected vehicles were fed into traffic applications such as max pressure signal control. Results showed that the cooperative perception environment could greatly augment the penetration rate in data collection. With 5% of CAVs and one infrastructure Lidar sensor, the E-CVPR can reach nearly 50%. The max pressure control application also showed comparable performances under both CAV and CV environments.

## ACKNOWLEDGEMENT

This research is supported in part by the U.S. National Science Foundation (NSF) through Grant CPS #2038215. The authors would like to thank Dr. Zhitong Huang from Leidos and the CARMA Program at FHWA for their technical support in the co-simulation environment. The views presented in this paper are those of the authors alone.

## AUTHOR CONTRIBUTION STATEMENT

The authors confirm their contributions to the paper as follows: study conception and design: Chen, Feng, Mao, Qian; data collection: Chen, Liu, Zhang; analysis and interpretation of results: Chen, Liu, Zhang, Feng; manuscript preparation: all authors. All authors reviewed the results and approved the final version of the manuscript.